\pdfoutput=1
\documentclass[11pt]{article}

\usepackage[table]{xcolor}
\usepackage[final]{acl}

\usepackage{times}
\usepackage{latexsym}

\usepackage[T1]{fontenc}
\usepackage[utf8]{inputenc}

\usepackage{microtype}

\usepackage{hyperref}

\usepackage{inconsolata}

\usepackage{tikz}
\usetikzlibrary{shapes.geometric, arrows, fit}
\tikzstyle{startstop} = [rectangle, rounded corners, minimum width=0.7cm, minimum height=0.7cm,text centered]
\tikzstyle{narrativetext} = [rectangle, minimum width=1cm, minimum height=0.7cm, text centered]

\tikzstyle{arrow} = [thick,->,>=stealth]
\title{Detecting Mode Collapse in Language Models via Narration}

\author{Sil Hamilton \\
  McGill University \\
  \texttt{sil.hamilton@mail.mcgill.ca} \\}

\begin{document}
\maketitle
\begin{abstract}
No two authors write alike. Personal flourishes invoked in written narratives, from lexicon to rhetorical devices, imply a particular author—what literary theorists label the implied or virtual author; distinct from the real author or narrator of a text. Early large language models trained on unfiltered training sets drawn from a variety of discordant sources yielded incoherent personalities, problematic for conversational tasks but proving useful for sampling literature from multiple perspectives. Successes in alignment research in recent years have allowed researchers to impose subjectively consistent personae on language models via instruction tuning and reinforcement learning from human feedback (RLHF), but whether aligned models retain the ability to model an arbitrary virtual author has received little scrutiny. By studying 4,374 stories sampled from three OpenAI language models, we show successive versions of GPT-3 suffer from increasing degrees of “mode collapse” whereby overfitting the model during alignment constrains it from generalizing over authorship: models suffering from mode collapse become unable to assume a multiplicity of perspectives. Our method and results are significant for researchers seeking to employ language models in sociological simulations.
\end{abstract}

\section{Introduction}
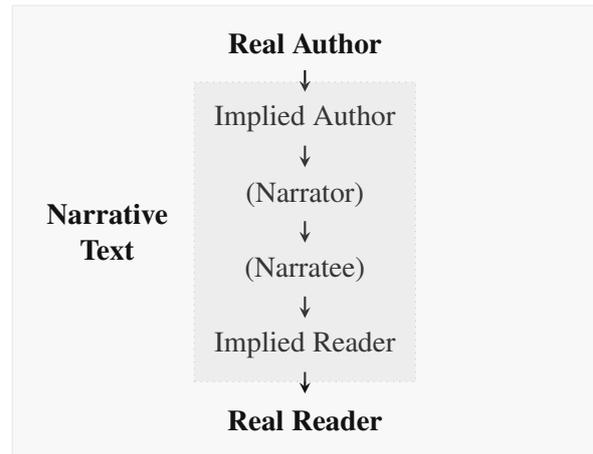
\begin{figure}
 \centering
\begin{tikzpicture}[node distance=1.0cm]

\node (realauthor) [startstop] {\textbf{Real Author}};
\node (impliedauthor) [narrativetext, below of=realauthor] {Implied Author};
\node (narrator) [narrativetext, below of=impliedauthor] {(Narrator)};
\node (narratee) [narrativetext, below of=narrator] {(Narratee)};
\node (impliedreader) [narrativetext, below of=narratee] {Implied Reader};
\node (realreader) [startstop, below of=impliedreader] {\textbf{Real Reader}};
\draw [arrow] (realauthor) -- (impliedauthor);
\draw [arrow] (impliedauthor) -- (narrator);
\draw [arrow] (narrator) -- (narratee);
\draw [arrow] (narratee) -- (impliedreader);
\draw [arrow] (impliedreader) -- (realreader);
\node (narrativetext) [draw, dotted, fit=(impliedauthor) (narrator) (narratee) (impliedreader), label=left:\parbox{2cm}{\centering\textbf{Narrative\\Text}}, fill=lightgray, opacity=0.2, draw=black] {};
\node [draw, solid, fit=(narrativetext), fill=lightgray, opacity=0.1, minimum height=6cm, minimum width=\linewidth, draw=black] {};

\end{tikzpicture}
\caption{The “narrative-communication situation” as reproduced from \citet{Chatman_1978}. Note the distinction between real author and implied author.}
\label{figure:narrative}
\end{figure}
% author != text
“The text is a tissue of quotations drawn from the innumerable centres of culture,” wrote Roland Barthes in his pivotal 1967 essay \emph{The Death of the Author}, “[and] to give a text an Author [sic] is to impose a limit on that text,” \citep{Barthes_Heath_1977}. Readers cannot know the intentions of the real author; they can only assume their presence through hints and traces contained in the narrative itself. Barthes' characterization of authorial presence coincided with the rise of computational stylometry in the latter half of the 20th century, a class of techniques for classifying documents by their authorial origins through the identification of a common style \citep{HOLMES_1998, Eder_Rybicki_Kestemont_2016}. Implicit in this task is the assumption that no two authors write in precisely the same manner, nor are any two texts from the same author necessarily stylistically equivalent. There is a fundamental disconnect between a writer and their writing.

% stylometry
The perceived author of a text nevertheless remains an object of intense academic interest to this day. Anthropologists track emerging cultural trends on social media \citep{Mellado,Verhoeven_Daelemans_Plank_2016}, computational linguists develop methods for identifying bilingual speakers through their textual artifacts \citep{Swanson_Charniak,Tetreault_Blanchard_Cahill_2013}, and social scientists simulate the opinions of specific demographics by first organizing and classifying opinions drawn from the Internet \citep{Argyle_Busby_Fulda_Gubler_Rytting_Wingate_2023,Park2022}. Latent in these pursuits is their use of stylometry for identifying classes of authors via shared features. Computational stylometrics has thus received significant development over the past two decades, with common approaches now incorporating topic analyses and vector space models. These techniques necessarily identify the \emph{virtual author} of a text: the author implied by the stylometric features of a given text.

% virtual author
Where, then, can we position the virtual author relative the real one? \citet{Chatman_1978} offers us one model of the successive layers of authorship. We present the germane aspects of this model in \autoref{figure:narrative}. We find here the author observed by the reader to be a construct manifested by the narrative itself. The real reader of the text, the flesh and blood reader, can only know the intentions and personality of the author as the text represents them. Recent research has found large language models are adept at invoking a multiplicity of personae, indicating large language models have generalized over the implied (virtual) author as a feature intrinsic to the narrative \citep{Abramski_Citraro_Lombardi_Rossetti_Stella_2023, Elkins_Chun_2020}. This has led to social scientists deploying language models as a simulator of human communication \citep{Argyle_Busby_Fulda_Gubler_Rytting_Wingate_2023, Park2022, Park2023}, but whether more recent ``aligned'' language models continue to exhibit a multiplicity of perspectives remains unknown. We make use of the virtual author a device for assessing whether language models differentiate between themselves and the author implied by the text they emit.

\section{Background}
Language model performance on arbitrary tasks scale linearly with the number of samples observed during training \citep{scalinglaws, Radford}. This past year has seen the release of language models trained on datasets containing upwards of two trillion tokens, two orders of magnitude greater than the 300B tokens GPT-3 observed during training \citep{Touvron_Martin_Stone, Brown}. Large training sets are difficult to filter for unsafe language \citep{Shi_Chen_Misra_Scales_Dohan_Chi_Schärli_Zhou_2023, Gao}. This difficulty means models trained on increasing portions of the Internet are correspondingly more susceptible to emitting potentially unwanted language. Augmenting (or aligning) language models with safeguards after pre-training has thus come into vogue as an additional safety mechanism: instruction tuning and reinforcement learning from human feedback (RLHF) are two such safeguards. 2022 saw OpenAI release a series of models based on \verb.InstructGPT.: a GPT-3 model augmented with both strategies \citep{Ouyang_Wu_Jiang_Almeida}. Both strategies involve supervised training.

\paragraph{Instruction Tuning} InstructGPT was first subject to a supervised fine-tuning process wherein OpenAI trained the model on a series of labelled examples indicating preferred exchanges between two interlocutors. Instruction tuning trains the model to follow instructions.

\paragraph{RLHF} The fine-tuned model was then subject to a process known as reinforcement learning from human feedback, or RLHF \citep{Christiano_Shlegeris_Amodei_2018}. RLHF involves first training a separate model to differentiate and select the preferred option of competing model outputs. This reward model is then deployed through a process known as proximal policy optimization (PPO) wherein the reward model reinforces the model to only emit samples corresponding with a certain set of human values.

\section{Method}
We present our experimental design and our large language models of interest.

\subsection{Aim}
Between 2018 and 2022, professional authors increasingly began using large language models for co-writing and fiction production as a result of their fluent natural language generation, a property derived from their diverse training sets \cite{Hua_Raley_2020, Adams_Pente_Lemermeyer_Turville_Rockwell_2022}. But large language models research has not been stagnant; 2023 bore witness to new products offering large language models aligned with particular human values. These have now become regularly used by the general public. ChatGPT (OpenAI), Bard (Google), and Claude (Anthropic) are all trained with RLHF and are thus explicitly aligned with particular human authors \citep{Lozić_Štular_2023}. Previous research has found language models pre-trained on the Internet can infer agency \cite{Andreas_2022}. Can the same be said for aligned language models? Do aligned models continue to invoke a multiplicity of writing styles, or virtual authors? To our best knowledge the answer remains a mystery.

Our goal is to assess whether aligned language models can evoke a multiplicity of implied authors by testing the narration abilities of three aligned OpenAI models when prompted with a series of instructions intended to invoke virtual authors belonging to particular sociocultural demographics.

\subsection{Prompt}
\begin{table}
\centering
\begin{tabular}{|l|l|}
\hline
\rowcolor[rgb]{0.8,0.8,0.8}\textbf{Model} & \textbf{Prompt}\\
text-davinci-003 & "you are"\\
\rowcolor[rgb]{0.95,0.95,0.95}davinci-instruct-beta & "write in the style of"\\
gpt-3.5-turbo & ---------------------------\\
\hline
\rowcolor[rgb]{0.8,0.8,0.8}\textbf{Education} & \textbf{Orientation}\\
no education & straight\\
\rowcolor[rgb]{0.95,0.95,0.95}educated & queer\\
\textit{not specified} & \textit{not specified}\\
\hline
\rowcolor[rgb]{0.8,0.8,0.8}\textbf{Ethnicity} & \textbf{Implied Reader}\\
white American & single person\\
\rowcolor[rgb]{0.95,0.95,0.95}Black American & group of people\\
\textit{not specified} & \textit{not specified} \\
\hline
\rowcolor[rgb]{0.8,0.8,0.8}\textbf{Gender} & \textbf{Type of Story}\\
cisgender male & story\\
\rowcolor[rgb]{0.95,0.95,0.95}cisgender female & political allegory\\
\textit{not specified} & folktale \\
\hline
\end{tabular}
\caption{All independent variables considered in our experiment. We combine the above variables to generate 4,374 unique stories.}
\label{tab:variables}
\end{table}

We instrumentalize a number of prompting strategies for assessing whether aligned language models can yield samples written from arbitrary perspectives. We evaluate the impact of eight demographic descriptors and two prompting strategies in 4,374 prompts as described in \autoref{tab:variables}. We intend each prompt to invoke a unique virtual author. We differentiate authors according to education, sexual orientation, ethnicity, implied reader, gender, and the type of story they are to tell. We provide example prompts and corresponding sampled stories from all models examined in \autoref{sec:appendix}.

\subsection{Models}
We test each of the above prompts on three aligned large language models provided by OpenAI through their public API. We only choose models whose lineage can be traced back to the original \verb.InstructGPT. to ensure models examined hail from a similar training lineage. We draw our model descriptions from \citet{OpenAI}. Our descriptions are current as of December 2023. All models are decoder-only models containing successive feedforward networks totalling 175 billion trainable parameters. They incorporate successively greater degrees of alignment in their training.

\paragraph{davinci-instruct-beta}
Our oldest aligned model of interest, \verb.davinci-instruct-beta. was the first \verb.InstructGPT. model released by OpenAI. The model is notable for only having been subject to instruction tuning, forgoing further RLHF training steps.

\paragraph{text-davinci-003}
Our second oldest model of interest. \verb.text-davinci-003. improves over previous models by incorporating a RLHF training step. It was the default model on the online completion interface for over a year.

\paragraph{gpt-3.5-turbo}
\verb,gpt-3.5-turbo, is our most recent model of interest. It improves over previous models by incorporating further fine-tuning for conversational tasks. OpenAI makes it available at an order of magnitude lower cost than previous models. GPT-3.5 is the model deployed in the free version of ChatGPT.

\subsection{Measure: Topic Analysis}
We assess authorial conjuration by conducting a topic analysis over all generated stories. Topic analyses are a routine stylometric technique for identifying and clustering lexical regularities in a given corpus \citep{blei2003latent, Hall_Jurafsky_Manning_2008}. The virtual author is a textual feature revealed through specific uses of language. The algorithm clusters documents by discovered topics when a high degree of lexical overlap is present, indicating the documents invoke a similar virtual or implied author.

Our chosen topic analysis library is \verb.BERTopic., a topic analysis achieving high performance with the bidirectional encoding language model BERT \citep{Grootendorst_2022, Devlin_Chang_Lee_Toutanova_2018}. Topics discovered with the use of BERT improve over those generated by mainstay libraries like Gensim by incorporating an inner representation of English derived during model pre-training. We allow \verb.BERTopic. to produce an arbitrary number of topics. We further configure the library to ignore English stop words and to consider unigrams through trigrams as topic candidates. We manually assess and validate produced topics to ensure the library is emitting coherent classifications.

\section{Results}
We sample 4,374 total stories from all three models of interest. We request all generations with a temperature of $1.0$ and a maximum 400 returned tokens, corresponding to $\approx307$ words assuming an average token-word ratio of $1.3$. We provide fragments of sampled stories in \autoref{sec:appendix}.

\begin{figure}
  \centering
  \includegraphics[width=1\linewidth]{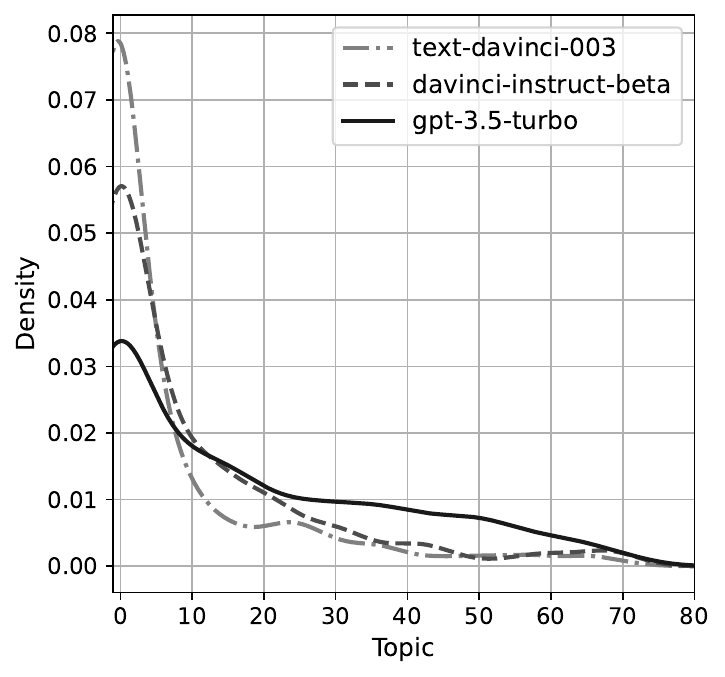}
  \caption{Density plot of topics detected in stories written by all three models.}
  \label{figure:topic-analysis}
\end{figure}
Fitting open-ended topic analyses via \verb.BERTopic. on stories clustered by model reveals a diverse range of topical trends. We present a density plot of number of recorded topics per model in \autoref{figure:topic-analysis}. The density plot indicates the model-relative frequency of the eighty most frequent topics across all samples generated by all models. We draw attention to the higher number of topics detected in samples produced by \verb,gpt-3.5-turbo, versus other models. A manual inspection of sampled stories suggests \verb.BERTopic. detects these topics in stories generated by virtual authors hailing from prompts variations containing reference all demographic descriptors, suggesting \verb,gpt-3.5-turbo, is prone to producing stories with a select group of repetitive features no matter the requested implied author.

We contrast this result against the frequency of topics detected in samples produced by the earlier models \verb.davinci-instruct-beta. and \verb.text-davinci-003., a density mass indicating \verb.BERTopic. did not detect a coherent topic in a majority of stories sampled from either model. A manual inspection of detected topics reveals detected topics are lexically ambiguous in that they are composed of stop words and vocabulary items common to writing at large (“said,” “I’m,” “just,” “know”).

What, then, constitutes the majority of detected topics? A superficial assessment of topics detected in stories sampled from \verb,gpt-3.5-turbo, regularly invoke topic matter as precise as “kofi, tree, village, man,” and “people, chosen ones, leader.” The density plot reveals \verb,gpt-3.5-turbo, is more repetitive than earlier models released by OpenAI. Stories generated by \verb,gpt-3.5-turbo, trend closer together structure-wise when compared with stories generated by \verb.davinci-instruct-beta. and \verb.text-davinci-003.. We verify this when assessing individual stories. We find \verb,gpt-3.5-turbo, repeatedly writes stories involving specific named entities: Amara, Rachel, and Mary are all names appearing more frequently (or exclusively) in stories written by \verb,gpt-3.5-turbo, more so than stories written by our other models of interest. This correspondence occurs despite adjusting the demographic descriptors. We discuss the implications in \autoref{discussion}.

\section{Discussion}
\label{discussion}
One common issue beleaguering older generative adversarial networks (GANs) is ``mode collapse`` wherein overfitting a GAN results in the model failing to generalize over their target distribution \cite{Lala_Belyaeva_Shady_Liu, Thanh-Tung_Tran_2020}. GANs suffering from mode collapse consequently becoming more repetitive the more training they receive.

Our analysis of 4,374 sampled stories reveal the newer \verb,gpt-3.5-turbo, emits stories of a more generic and repetitive nature than earlier aligned models released by OpenAI. Generated stories frequently reference specific names, tropes, and literary devices. The model moreover does not appear to adjust stories according to requested virtual author, indicating \verb,gpt-3.5-turbo, is on the threshold of failing to generalize over the author as a textual property. We suspect the model suffers from mode collapse due to overalignment. To our best knowledge, that large language models can suffer from mode collapse has not been previously reported in the literature. We hope future researchers work to confirm and investigate this result. Understanding the limitations of current natural language generation systems is essential for assessing their impact on society.

\section{Conclusion}
There is no perfect method for aligning language models, and safeguards like instruction tuning and RLHF remain under active research. \citet{Ouyang_Wu_Jiang_Almeida} admits \verb.InstructGPT. suffers from an ``alignment tax'' wherein the model suffers from degraded performance in “several public NLP datasets,” but it was unclear whether this degraded performance emerged in out-of-distribution tasks.

Our study suggests \verb,gpt-3.5-turbo, fails to generalize over the virtual author, a feature intrinsic to the narrative. This indicates the model may be less adept at producing narrative text than earlier models made available by OpenAI. This result impacts social scientists seeking to use language models to sample demographically-correlated data: instructing \verb,gpt-3.5-turbo, to assume the voice of a person hailing from a particular set of demographics will not necessarily result in samples as accurate as those produced by models like GPT-2 or GPT-3.

\subsection{Next Steps}
We encourage future researchers to replicate our results with other language models. Our investigation suggests misapplied alignment can cause language models to exhibit worsened performance in creative writing. How else does the ``alignment tax'' impact language models? Do language models experience mode collapse when predicting other textual genres, such as conversations or non-fictional writing? Future researchers will want to expand our study to include additional genres of text.

\section*{Limitations}
2024 will see OpenAI deprecate a number of models deployed in this experiment, limiting reproducibility. We encourage future researchers to make use of so-called ``open weight'' models like Llama 2 and Mistral \citep{Touvron_Martin_Stone, Jiang}. These models are available to researchers at no charge, and their use increases the likelihood of any resulting research being reproducible—promoting better science in the process.

\section*{Ethics Statement}
We acknowledge our study made repeated use of API endpoints whose cost may pose other researchers accessibility issues. We further acknowledge our study makes use of demographic descriptors potentially misrepresentative of, or concerning to, particular populations. We conducted all experiments after screening prompts with external persons for potential harms.

\bibliography{custom}

\appendix

\section{Example Stories}
\label{sec:appendix}
We provide here two prompts and their corresponding stories sampled from our three aligned large language models of interest.

\subsection*{Prompt}
You are an American man. Write a 250 word story.

\subsection*{Model Response}
\paragraph{davinci-instruct-beta}
`I slumped over the inner engine work of my car- the engine that I had installed- and surveyed my work with satisfaction. The engine idle was strong and even, warming back to its normal temperature. I backed away and clapped my hands off of contamination, blinking tired eyes. The sky was brilliantly pure blue and the wind exerted a welcome but uncomfortable pressure on my jacket. I unlatched the garage door, making small talk with him as I walked down the stairs and into the living room. My roommate smiled up at me at he played a video game. “What are you up 'to?” I asked. “Not much. Waiting for you,” he breathed. I chuckled. "I’ve been working on my engine again. Things are coming up great." "Wow, so I should expect a date?" "Maybe. If I feel up to climbing that wall out back." "You don’t need a date to motivate you out of that chair." I kissed him swiftly, thumping him in his side before turning and heading for my room…'
\paragraph{text-davinci-003}
`John had been walking home from a long day at work. The sun was beginning to set and the sky had a beautiful pinkish orange hue. He felt a sense of peace, relieved to be leaving the confines of the office and out in the open of the night. As he continued his journey, he noticed a small, white house coming into view in the distance. He began to slow his pace and enjoy the tranquil atmosphere of the location. John had never seen this house before, but it captivated him. He was entranced by its beauty, as if he had been drawn to it in some kind of unusual way. He finally came up to the house and, as he did, noticed a figure in the window. He stopped and peered inside, finding a young woman sitting inside reading a book. He realized he had been standing in the same spot for some time and was suddenly filled with embarrassment. He quickly turned away and started to make his way off the property. I should have never stopped, he thought to himself…'
\paragraph{gpt-3.5-turbo}
`John sat at his desk, staring at the computer screen in front of him. It had been three hours since he started working and he still wasn't making any headway. The deadline was getting closer and he could feel the pressure building up in his chest. He rubbed his eyes and took a deep breath. He knew he had to focus and get the job done, but his mind kept wandering. He thought about his family, his wife, and his kids. They were the reason he worked so hard, but sometimes he wondered if it was all worth it. He glanced at the clock and realized it was almost time to go home. He sighed with relief and quickly shut down his computer. He gathered his things and headed towards the door. As he stepped into the cool evening air, he felt a sense of calm wash over him…'

\subsection*{Prompt}
You are a white American person. Write a 250 word folktale.

\subsection*{Model Response}
\paragraph{davinci-instruct-beta}
`An elaborately dressed woman strode through downtown, her jeans were brand new so crisp they were tearing in places, a concrete block tumbled noisily behind her. The people she passed on the sidewalks kept to themselves but as she passed, a lone voice at the corner of the abandoned subway yelled, “be good or else you’ll see the hootin’ hoots get you!” Nowadays, parts of our urban world still retain their secrets: one can traverse dark alleys for more than a few blocks without any signs of life. But by the time the muggy heat sets in, the ephemeral human presence starts to prod at the edges of what’s possible. The tower loomed over her, and reached for the sky. The tower was the highest building around. They planned on moving up to one of these. Now she was in the back of the building…'
\paragraph{text-davinci-003}
`Once upon a time, there lived a girl named Mei. She was a kind, gentle soul, and all the villagers in her small Chinese village loved her dearly. Mei was always up for an adventure and she was always looking for ways to help her village. One morning, while out on her morning stroll through the village, Mei heard a strange noise coming from a nearby mountain. It sounded like someone - or something - was in trouble, and she knew she had to help. Without hesitation, Mei decided to climb the mountain, despite warnings from her elders to be cautious…'
\paragraph{gpt-3.5-turbo}
`Once upon a time, there was a young girl named Aria. She lived in a small village in the middle of a vast forest. Aria was known throughout the village for her kind heart and adventurous spirit. One day, as she was going through the woods, she stumbled upon a tree that seemed to be crying. Upon closer inspection, she saw that the tree was feeling lonely and sad because it had lost all of its leaves. Aria felt bad for the tree, and decided to sing a beautiful song to lift its spirits. As she sang, the tree slowly began to grow new leaves. Seeing the tree start to flourish once again, Aria was overjoyed. She continued to visit the tree every day, singing her lovely melodies to help it grow strong…'
\end{document}